\documentclass{article}
\usepackage{spconf,amsmath,graphicx,hyperref}
\usepackage{amsfonts}
\usepackage{multirow}
\usepackage{graphicx}
\usepackage{subcaption}
\usepackage{tabularx}
\usepackage{hyperref}
\usepackage{booktabs}
\usepackage{xcolor}

\title{Multi-Scale Adaptive Neighborhood Awareness Transformer For Graph Fraud Detection}
\name{Jiaqi Lv, Qingfeng Du$^\ast$\thanks{$^\ast$Corresponding author: Qingfeng Du. This work was supported by National Science and Technology Major Project of China (Grant No. 2022ZD0120303).}, Yu Zhang, Yongqi Han, Sheng Li
}
\address{School of Computer Science and Technology, Tongji University, Shanghai, China}

\begin{document}
%
\maketitle
\begin{abstract}
Graph fraud detection (GFD) is crucial for identifying fraudulent behavior within graphs, benefiting various domains such as financial networks and social media. Existing methods based on graph neural networks (GNNs) have succeeded considerably due to their effective expressive capacity for graph-structured data. However, the inherent inductive bias of GNNs, including the homogeneity assumption and the limited global modeling ability,  hinder the effectiveness of these models. To address these challenges, we propose Multi-scale Neighborhood Awareness Transformer (MANDATE), which alleviates the inherent inductive bias of GNNs. Specifically, we design a multi-scale positional encoding strategy to encode the positional information of various distances from the central node. By incorporating it with the self-attention mechanism, the global modeling ability can be enhanced significantly. Meanwhile, we design different embedding strategies for homophilic and heterophilic connections. This mitigates the homophily distribution differences between benign and fraudulent nodes. Moreover, an embedding fusion strategy is designed for multi-relation graphs, which alleviates the distribution bias caused by different relationships. Experiments on three fraud detection datasets demonstrate the superiority of MANDATE.
\end{abstract}

\section{Introduction}

With the rapid development of the information technology industry, fraudulent activities have increased significantly in financial networks \cite{financial_networks}, social media \cite{social_media}, review networks \cite{review_networks}, and e-Commerce platforms \cite{you2024multi}. These activities not only result in severe financial losses for companies, but also threaten the safety and privacy of consumers. Consequently, the detection of such activities has become an area of paramount importance.

As fraud typically manifests within abnormal interactions between entities, recent research has focused primarily on graph-based fraud detection (GFD). Meanwhile, Graph Neural Networks (GNNs) have shown great potential in GFD due to their effective expressive capacity for graph-structured data. However, the inherent inductive bias in GNNs can hinder fraud detection performance in two ways: (1) the homogeneity assumption \cite{gao2025homogeneous} and (2) the limited global modeling ability. To mitigate the performance degradation caused by the homogeneity assumption, a promising approach is to reduce the heterophily distribution in GFD. GHRN \cite{GHRN} and GDN use high-pass filters to selectively remove heterophilic edges. However, these irreversible methods can disrupt the structural information of the graph to some extent. H2-FDetector \cite{H2-FDetector} makes homophilic connections propagate similar information and heterophilic connections propagate different information. This assumption may not always be valid in practical scenarios.

Thus, we propose MANDATE (Multi-scale Adaptive Neighborhood Awareness Transformer), which alleviates the inherent inductive bias of GNNs. First, we design a multi-scale positional encoding strategy to capture positional information more extensively. Concretely, MANDATE encodes the positional information of different distances from the central node and dynamically adjusts the attention to these encodings. Second, we present a neighborhood awareness positional embedding module to extract the positional embeddings. Different embedding strategies are designed for homophilic and heterophilic connections. This mitigates the homophily distribution differences between benign and fraudulent nodes. Moreover, we design an embedding fusion strategy for multi-relation graphs, alleviating the distribution bias caused by different relationships. 

\section{Related Work}

Graph Fraud Detection (GFD) is a critical problem in graph mining. Several GNN-based detectors have been proposed to enhance the exploration of fraud patterns through graph structures and node attributes. These detectors can be divided into spectral-based and spatial-based according to the information propagation method. Spectral-based methods define graph convolution operations through eigenvalue decomposition of the graph's Laplacian matrix. GWNN \cite{GWNN} leverages graph wavelet transform to address the shortcomings of previous spectral graph CNN methods that depend on the graph Fourier transform. On the other hand, GraphSAGE \cite{GraphSAGE} and GIN \cite{GIN} explore from the perspective of neighbor aggregation. CARE-GNN \cite{CARE-GNN} and PC-GNN \cite{PC-GNN} further provide a similarity-based neighbor selection strategy through threshold filtering. Despite the progress made by existing approaches, the homogeneity assumption of GNN-based models obscures critical distinctions between fraudulent and normal nodes. In addition, local message-passing mechanisms limit their capacity to model global fraud patterns across communities.

\begin{figure*}
    \centering
    \includegraphics[width=0.93\linewidth]{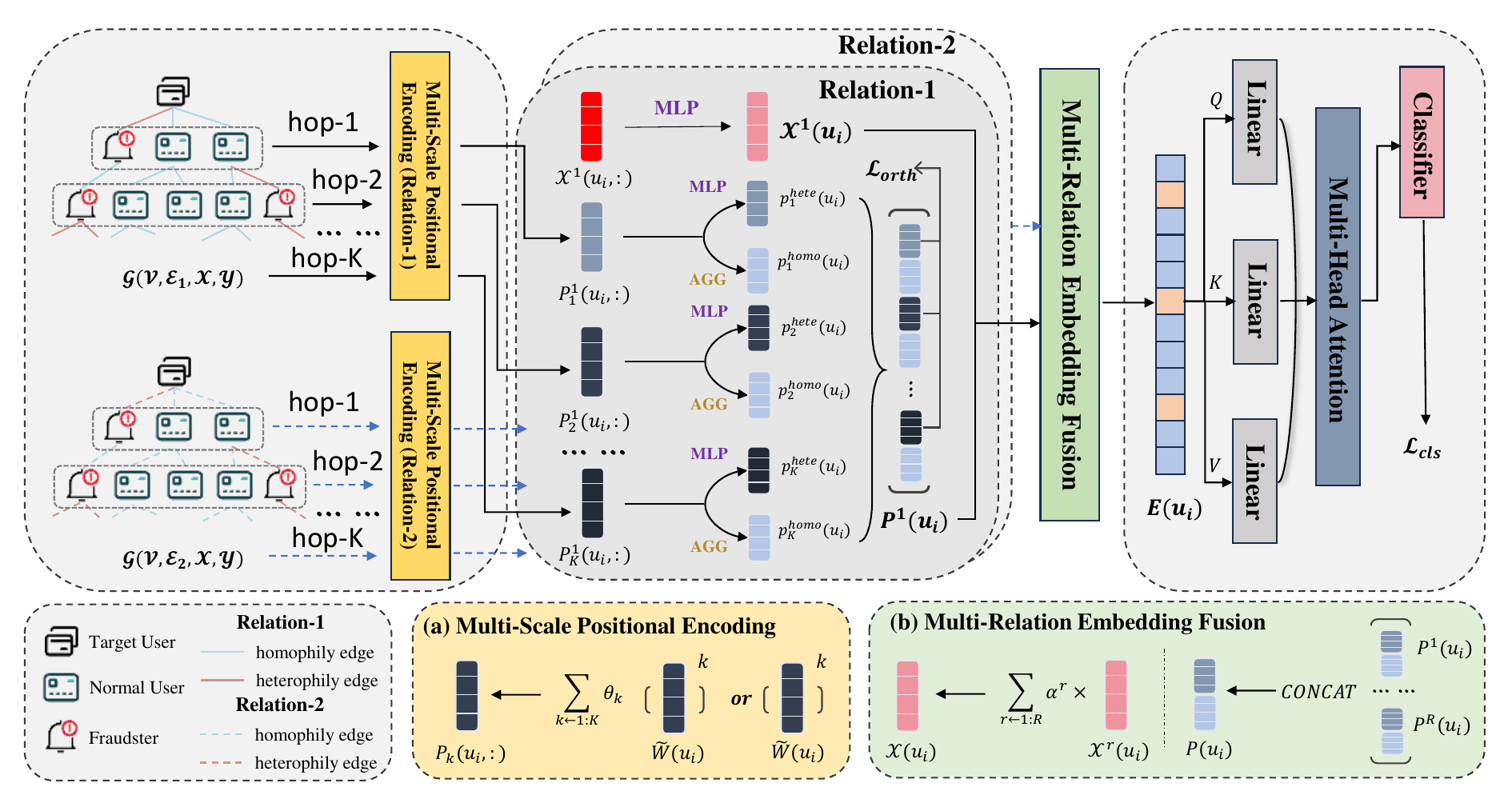}
    \caption{The overall architecture of MANDATE. For simplicity, we set the number of relationships to 2 as an example. }
    \label{fig:framework}
\end{figure*}

\section{Methodology}

\subsection{Multi-scale Positional Encoding Strategy}
\label{Multi-scale Positional Encoding Strategy}

As shown in Figure \ref{fig:framework}, the positional encoding of node $u_i$ can be denoted as $p(u_i)$, which is used to encode positional information $\phi(l_{u_i,u_j})$ from $u_i$ to any $u_j \in \mathcal{V}$. Here, $\phi(\cdot)$ is an encoding function and $l_{u_i,u_j}$ is the positional information. $\phi(\cdot)$ can be implemented with various encoding strategies, such as the shortest path distance or generalized PageRank scores. To capture multi-scale positional information, we set $p(u_i)$ to a sequence $S_{p(u_i)}$.

\begin{equation}
    S_{p(u_i)} = \left \{ p_k\left ( u_i \right ) \mid k= 1,2,\dots ,K \right \}
\end{equation}

$p_k(u_i)$ corresponds to the positional encoding of \textit{k}-hop. We first exploit random walk matrix to represent positional information. $\phi(\cdot)$ represents the shortest path distance, while $\phi_k( l_{u_i,u_j})$ corresponds to a \textit{k}-hop random walk.

\begin{equation}
    \phi_k( l_{u_i,u_j}) = \tilde{W}^k(u_i,u_j) = (D^{-1} A)^k(u_i,u_j)
\end{equation}

Here, $A$ is the adjacency matrix, $D$ is the respective degree matrix. We represent the \textit{i}-th row of matrix $\tilde{W}$ as $\tilde{W}(u_i,:)$ and the \textit{(u,v)}-th entry as $\tilde{W}(u_i,u_j)$.

\begin{equation}
    p_k(u_i) = \{\phi_k( l_{u_i,u_j}) | u_j \in \mathcal{V}\} = \tilde{W}^k(u_i,:)
    \label{eq:positional_encoding}
\end{equation}

Theoretically, $S_{p(u_i)}$ can capture all the shortest path distances for nodes up to \textit{K} hops away from each other. In particular, when we set \textit{K} to n, we can capture all the shortest path distances and mark unreachable nodes. If we learn new end-to-end positional encoding with \textit{MLP}, $S_{p(u_i)}$ can also approximate general classes of graph propagation matrices $p^{'}(u_i)$ through learnable parameters.

\begin{equation}
    p^{'}(u_i) = MLP(S_{p(u_i)}) = \sum_{k=1}^{K} \theta_k \tilde{W}^k(u_i,:)
\end{equation}

$\theta_k$ is the learnable parameter for hop k. If $K=1$, $p^{'}(u_i)$ can represent weighted neighbor aggregation information. While it can denote generalized PageRank scores when $\theta_k = \alpha(1-\alpha)$. Such graph propagation matrices can provide more comprehensive positional information. When introducing the proposed positional embedding in the next subsection, we will also treat $p^{'}(u_i)$ as a part of the positional embedding.

\subsection{Neighborhood Awareness Positional Embedding}
\label{Neighborhood Awareness Positional Embedding}

For homophilic connections, neighboring nodes tend to have similar attributes and labels. $\phi_k( l_{u_i,u_j})$ measures the distance between node $u_i$ and node $u_j$. Therefore, we can directly employ $\phi_k( l_{u_i,u_j})$ as the probability of propagating feature from $u_i$ to $u_j$.

\begin{equation}
    p_k^{homo}(u_i) = \sum_{u_j \in \mathcal{V}}^{} \phi_k( l_{u_i,u_j}) \mathcal{X}(u_j,:) 
\end{equation}

Heterophilic connections refer to relationships in which topological closeness does not mean positive correlations. Consequently, unlike in the case of homophilic connections, it is not feasible to directly aggregate neighbor information to the central node. Therefore, we exploit a neural network architecture to automatically learn such complex correlations. Specifically, we concatenate positional information $p_k(u_i)$ and attribute information $\mathcal{X}(u_i,:)$. Then, we feed it to the neural network. Mathematically, positional embeddings can be computed as follows:

\begin{equation}
    p_k^{hete}(u_i) = MLP(CONCAT(p_k(u_i),\mathcal{X}(u_i,:))
\end{equation}

We combine the embedding information from these two perspectives. The \textit{k}-hop positional embedding can be presented as:

\begin{equation}
    p_k^{'}(u_i) = p_k^{homo}(u_i) || p_k^{hete}(u_i)
\end{equation}

Finally, the \textit{k}-hop positional embeddings are concatenated and subjected to \textit{MLP} to generate the final positional embedding. Here, \textit{MLP} allows the model to learn the most appropriate range of positional information.

\begin{equation}
    P(u_i) = MLP(CONCAT(p_1^{'}(u_i),\dots,p_K^{'}(u_i),p^{'}(u_i)))
\end{equation}

\begin{equation}
    E(u_i) = CONCAT(\mathcal{X}(u_i,:),P(u_i))
\end{equation}

$\mathcal{X}(u_i,:)$ is the node-specific features, $P(u_i)$ is the positional features, and $E(u_i)$ represents the final embedding of node $u_i$. Moreover, there is information redundancy between the \textit{k}-hop positional embeddings. High-order embeddings inherently contain some of the information from low-order embeddings. To preserve the unique positional information for embeddings of different order, we impose orthogonality constraints on the positional information. Concretely, we apply a cosine embedding loss that directly minimizes angular similarities of these embeddings.

\begin{equation}
    \mathcal{L}_{orth} = \sum_{1 \leq m < n \leq K}^{}  \left \langle p_m^{'}(u_i),p_n^{'}(u_i) \right \rangle 
    \label{eq:L_orth}
\end{equation}

\subsection{Multi-relation Embedding Fusion Strategy}
\label{Multi-relation Embedding Fusion Strategy}

For node-specific embeddings, we denote it as $F(u_i)$ and aggregate all embeddings by using a set of learnable weight parameters as follows:

\begin{equation}
    F^{'}(u_i) = \sum_{r=1}^{R} \alpha ^r F^{r}(u_i)
\end{equation}

This can be viewed as using a multi-head network to learn different attention representations in parallel, thereby providing more diverse representations. For positional embeddings, since the homophily distributions of different relations can obtain different aspects of information, these positional embeddings are equally significant. We concatenate the positional embeddings of different relations and allow the model to autonomously learn the influence of each relation. The embeddings of node $u_i$ for multi-relation graphs is:

\begin{equation}
    E(u_i) = CONCAT(F^{'}(u_i),P^{1}(u_i),\dots,P^{R}(u_i))
\end{equation}

\section{Experiments}

\begin{table*}[t]
\caption{Experimental results of all compared methods on YelpChi, Amazon and T-Finance. The best result is shown in \textbf{bold}, while the second best is marked with \underline{underline}. OOM means out of memory.}
\centering
  \begin{tabular}{r|ccc|ccc|ccc}
    \toprule
    Dataset& \multicolumn{3}{c|}{YelpChi} & \multicolumn{3}{c|}{Amazon} & \multicolumn{3}{c}{T-Finance}\\
    \cmidrule(lr){2-4}
    \cmidrule(lr){5-7}
    \cmidrule(lr){8-10}
    Metric & AUC & F1-macro & Gmean & AUC & F1-macro & Gmean & AUC & F1-macro & Gmean\\ \midrule
    GraphSAGE{\scriptsize\textsubscript{(NIPS'17)}}   & 0.8087 & 0.7161 & 0.7338 & 0.9401 & 0.9191 & 0.8949 & 0.9270 & 0.9083 & 0.8893\\
    CARE-GNN{\scriptsize\textsubscript{(CIKM'20)}}    & 0.7720 & 0.6116 & 0.7060 & 0.9041 & 0.8644 & 0.8572 & 0.9169 & 0.7522 & 0.8403\\
    DiG-in-GNN{\scriptsize\textsubscript{(AAAI'24)}}  & 0.9313 & 0.8136 & 0.8209 & \underline{0.9783} & 0.9259 & 0.9113 & 0.9642 & 0.9140 & \underline{0.8956}\\ \midrule
    AMNet{\scriptsize\textsubscript{(IJCAI'22)}}      & 0.8364 & 0.6780 & 0.5656 & 0.9642 & 0.9147 & 0.8863 & 0.9295 & 0.8672 & 0.7835\\
    BWGNN{\scriptsize\textsubscript{(ICML'22)}}      & 0.9119 & 0.7712 & 0.8022 & 0.9755 & 0.9249 & 0.8934 & 0.9583 & 0.8925 & 0.8435\\
    GHRN{\scriptsize\textsubscript{(WWW'23)}}         & 0.9042 & 0.7700 & 0.7838 & 0.9746 & 0.9257 & 0.8939 & 0.9606 & 0.8905 & 0.8325\\ \midrule
    H2-FDetector{\scriptsize\textsubscript{(WWW'22)}} & 0.8962 & 0.7147 & 0.8224 & 0.9682 & 0.8008 & 0.9173 & 0.9587 & 0.6097 & 0.8719\\
    GTAN{\scriptsize\textsubscript{(AAAI'23)}}        & 0.9038 & 0.7733 & 0.6999 & 0.9342 & 0.9103 & 0.8656 & OOM   & OOM    & OOM   \\
    ConsisGAD{\scriptsize\textsubscript{(ICLR'24)}}   & 0.8916 & 0.7514 & 0.7442 & 0.9735 & \underline{0.9276} & \underline{0.9232} & \textbf{0.9738} & 0.9121 & 0.8741\\
    PMP{\scriptsize\textsubscript{(ICLR'24)}}         & \underline{0.9320} & \underline{0.8140} & \underline{0.8402} & 0.9759 & 0.9088 & 0.9089 & 0.9680 & \underline{0.9202} & 0.8815\\ \midrule
    MANDATE (ours)  & \textbf{0.9473} & \textbf{0.8335} & \textbf{0.8607} & \textbf{0.9871} & \textbf{0.9280} & \textbf{0.9331} &        \underline{0.9712}  & \textbf{0.9222} &       \textbf{0.9203} \\ 
    \bottomrule
\end{tabular}
\label{tab:overall_exp}
\end{table*}

\subsection{Experimental Setup}

\textbf{Datasets}. We conduct experiments on three widely used fraud detection datasets: YelpChi, Amazon, and T-Finance. YelpChi is collected from Yelp.com. Amazon is based on user review data from Amazon.com. T-Finance is designed to detect anomalous users within a financial network.

\textbf{Implementation Details}. The implementation of MANDATE is based on PyTorch, and all experiments are conducted on an NVIDIA Tesla L40 48GB GPU. Following \cite{DiG-In-GNN}, we adopt a data split ratio of 40\% for training, 20\% for validation, and 40\% for the test set. Early stopping is applied if there is no improvement in validation performance within 20 epochs. 

\textbf{Comparison Methods.} We compared MANDATE with several state-of-the-art fraud detection methods to evaluate its effectiveness. We also choose to use GraphSAGE \cite{GraphSAGE} as a baseline for the GNN-based models. \textbf{(1) Neighbor Selection Based Methods:} CARE-GNN \cite{CARE-GNN}, and DiG-in-GNN \cite{DiG-In-GNN}. \textbf{(2) Frequency-Optimized Based Methods:} AMNet \cite{AMNet}, BWGNN \cite{T_finance_BW_GNN}, and GHRN \cite{GHRN}. \textbf{(3) Aggregation Strategy-Optimized Based Methods:} H2-FDetector \cite{H2-FDetector}, GTAN \cite{GTAN}, ConsisGAD \cite{ConsisGAD}, and PMP \cite{PMP}. 

\subsection{Fraud Detection Performance}

\begin{figure}[htbp]
    \centering
    \begin{minipage}{0.235\textwidth}
        \centering
        \includegraphics[width=\linewidth]{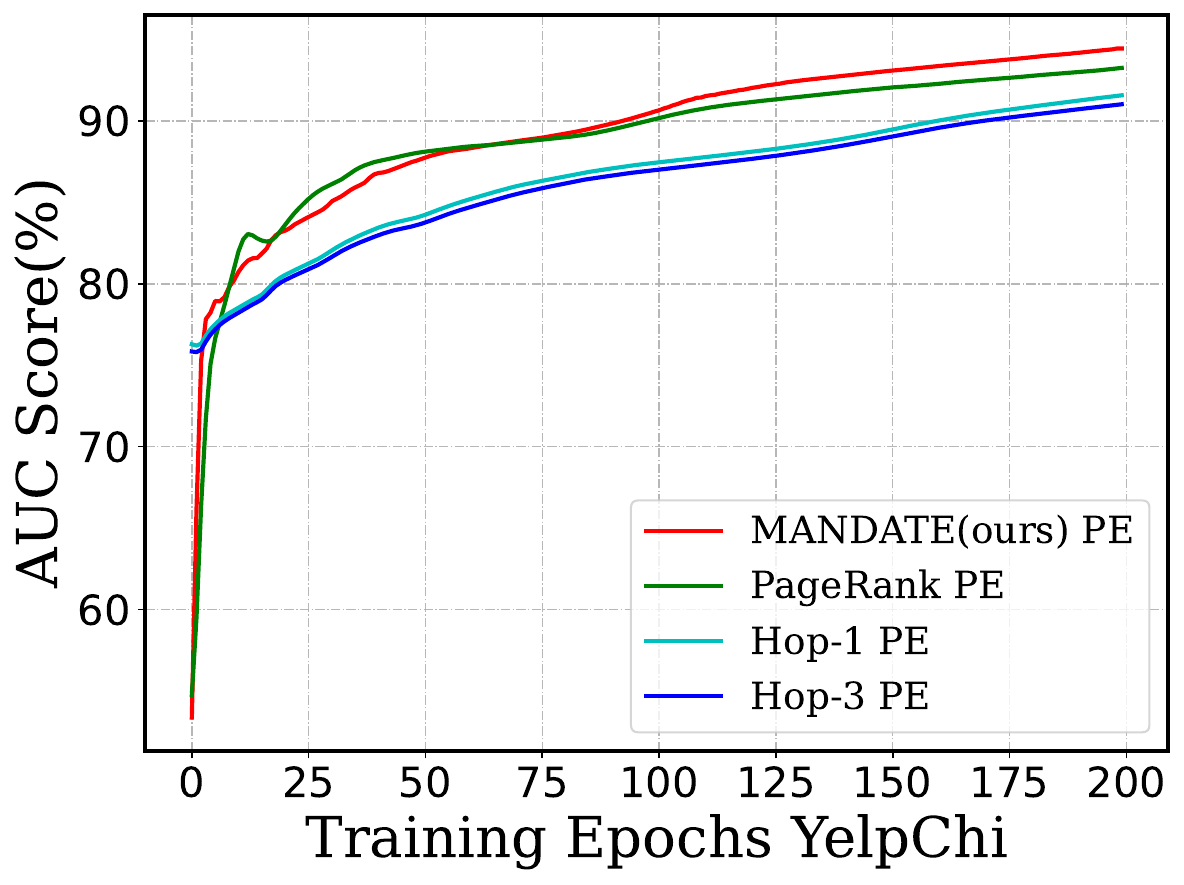}
        \label{fig:yelpchi_line_chart}
    \end{minipage}
    \begin{minipage}{0.235\textwidth}
        \centering
        \includegraphics[width=\linewidth]{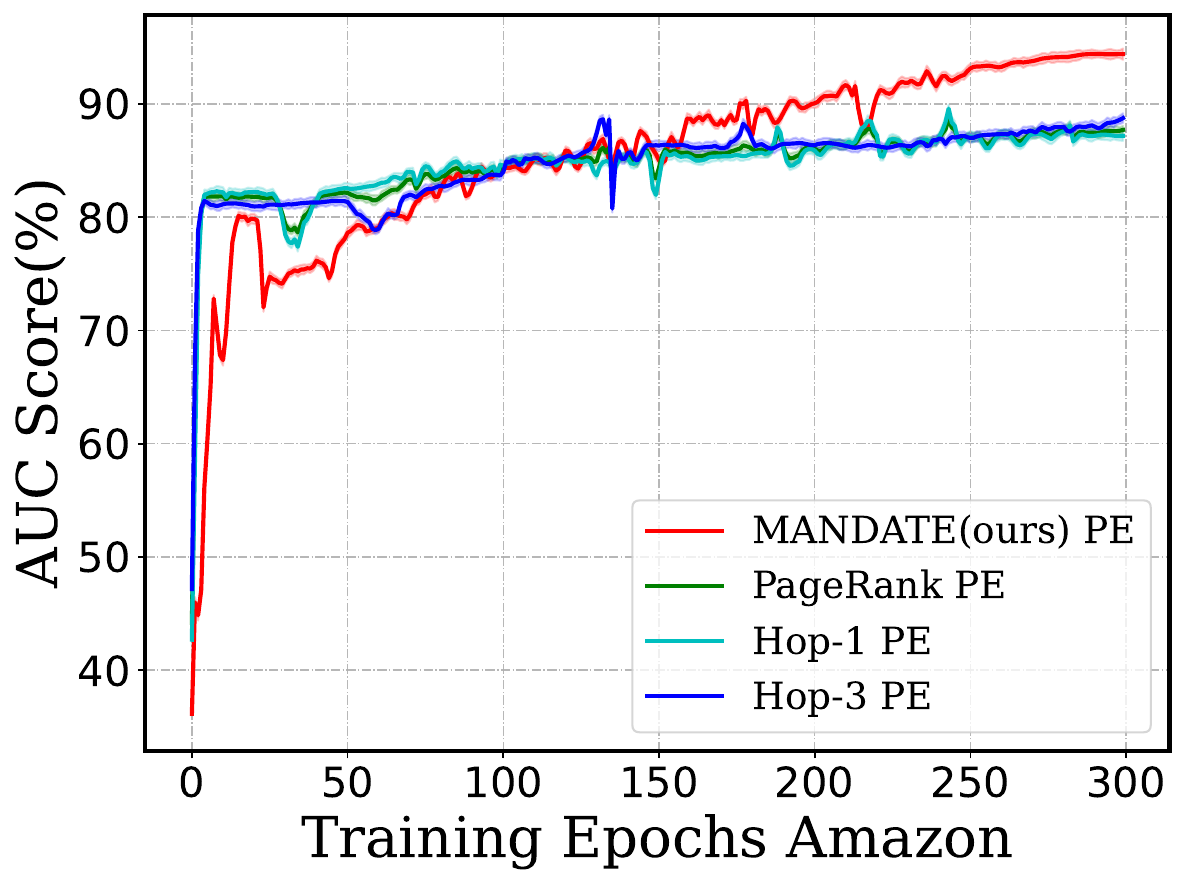}
        \label{fig:amazon_line_chart}
    \end{minipage}
    \caption{Performance comparison of MANDATE under the diverse positional encoding strategy on the YelpChi and Amazon datasets.}
    \label{fig:pe}
\end{figure}

In general, the following observations presented in Table~\ref{tab:overall_exp} demonstrate that our proposed MANDATE method achieves the best performance across all three datasets. On the YelpChi dataset, MANDATE outperforms other methods by up to 17.04\% on the F1-macro and Gmean metrics, respectively, with an improvement in AUC as well. Similarly, on the Amazon and T-Finance datasets, MANDATE exhibits comparable advantages, confirming the widespread effectiveness and strong competitiveness of our approach across different datasets.

\subsection{Ablation Study}

\begin{figure}[htbp]
    \centering
    \begin{minipage}{0.235\textwidth}
        \centering
        \includegraphics[width=\linewidth]{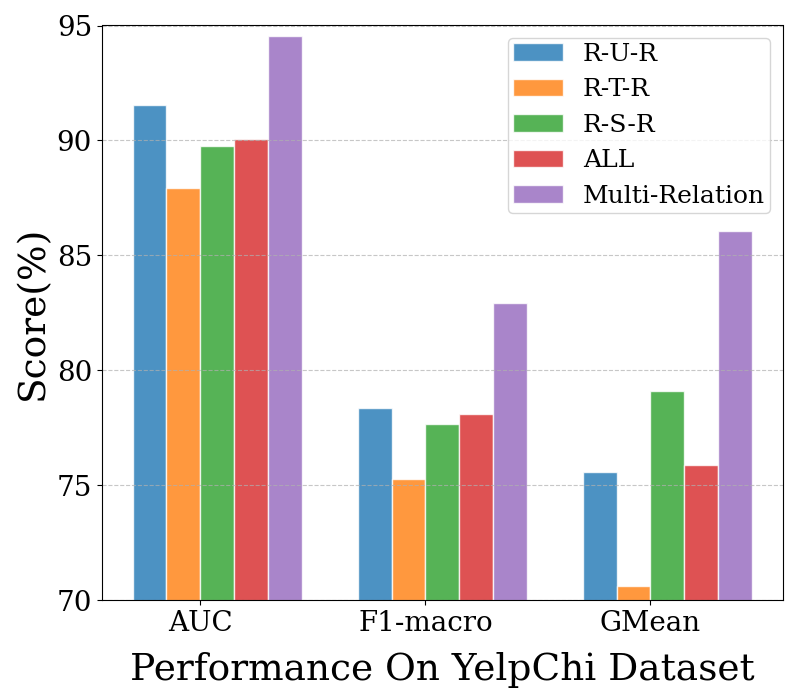}
        \label{fig:yelpchi_hist_chart}
    \end{minipage}
    \begin{minipage}{0.235\textwidth}
        \centering
        \includegraphics[width=\linewidth]{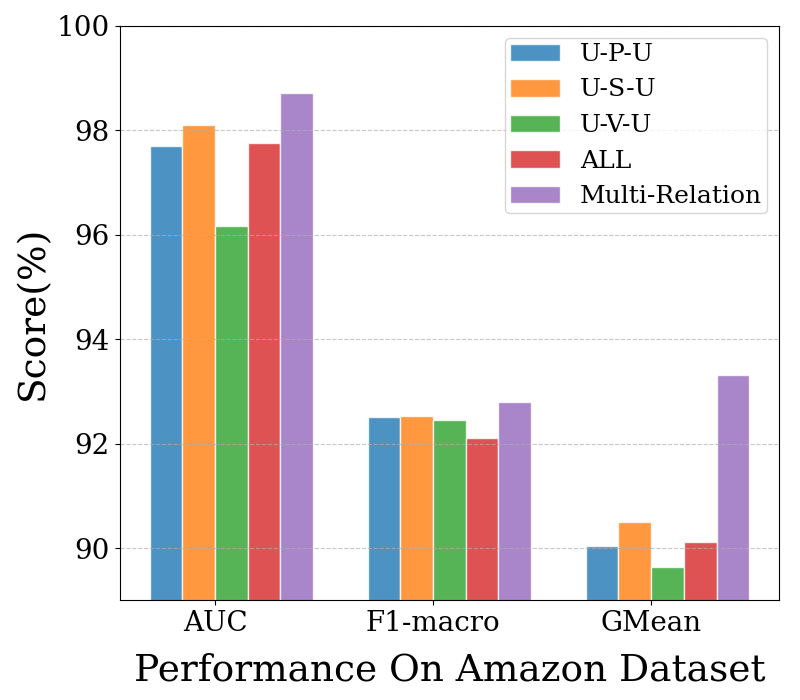}
        \label{fig:amazon_hist_chart}
    \end{minipage}
    \caption{Performance comparison of MANDATE under the multi-relation fusion strategy on the YelpChi and Amazon datasets.}
    \label{fig:multi_realtion}
\end{figure}

To assess the effectiveness of the proposed multi-scale positional encoding strategy, we compare it with PageRank and single-hop encoding strategies. As shown in Figure \ref{fig:pe}, our encoding strategy is generally better than other methods. Furthermore, we conduct experiments on the YelpChi and Amazon datasets to assess the effectiveness of multi-relation embedding fusion by comparing MANDATE using a single relation with its multi-relation fusion variant. As illustrated in Figure \ref{fig:multi_realtion}, MANDATE with multi-relation embedding fusion significantly outperforms its single-relation counterparts in fraud detection. Specifically, on the structurally more complex YelpChi dataset, the multi-relation embedding fusion strategy boosts AUC by at least 3\% over the single-relation approach, with even more pronounced improvements in Gmean and F1-Macro. 

\section{Conclusion}

The homogeneity assumption and the limited global modeling ability hinder the effectiveness of GNNs. Moreover, the intrinsic homophily distribution differences within GFD tasks further enhance this performance degradation. In this paper, we present a multi-scale positional encoding and neighborhood awareness positional embedding strategy. Meanwhile, we design a relation embedding fusion strategy for multi-relation graphs. MANDATE alleviates the inherent inductive bias of GNNs and enhances the compatibility of Transformers with GFD tasks. 

\bibliographystyle{IEEEbib}
\bibliography{strings,refs}

\end{document}